\documentclass[11pt,a4paper]{article}
\usepackage[hyperref]{naaclhlt2019}
\usepackage{times}
\usepackage{latexsym}
\usepackage{bm}
\usepackage{graphics}
\usepackage{graphicx}
\usepackage{epstopdf}
\usepackage{multirow}
\usepackage{threeparttable}
\usepackage{amsfonts}
\usepackage{stfloats}

\usepackage{url}

\aclfinalcopy 

\begin{document}

\author{Yu Cao$^1$~~~~Meng Fang$^2$~~~~Dacheng Tao$^1$ \\
$^1$UBTECH Sydney AI Center, School of Computer Science, FEIT, The University of Sydney, Australia \\
$^2$Tencent Robotics X, China \\
\tt ycao8647@uni.sydney.edu.au, mfang@tencent.com,\\ \tt dacheng.tao@sydney.edu.au}
  
\title{BAG: Bi-directional Attention Entity Graph Convolutional Network for Multi-hop Reasoning Question Answering}

\maketitle
\begin{abstract}
Multi-hop reasoning question answering requires deep comprehension of relationships between various documents and queries. We propose a Bi-directional Attention Entity Graph Convolutional Network (BAG), 
leveraging relationships between nodes in an entity graph and attention information between a query and the entity graph, to solve this task.
Graph convolutional networks are used to obtain a relation-aware representation of nodes for entity graphs built from documents with multi-level features. 
Bidirectional attention is then applied on graphs and queries to generate a query-aware nodes representation, which will be used for the final prediction. Experimental evaluation shows BAG achieves state-of-the-art accuracy performance on the QAngaroo WIKIHOP dataset.
\end{abstract}

\section{Introduction}

Question Answering (QA) and Machine Comprehension (MC) tasks have drawn significant attention during the past years. The proposal of large-scale single-document-based QA/MC  datasets, such as SQuAD~\cite{squad}, CNN/Daily mail~\cite{cnndaily}, makes training available for end-to-end deep neural models, such as BiDAF~\cite{bidaf}, DCN~\cite{dcn} and SAN~\cite{san}. However, gaps still exist between these datasets and real-world applications. For example, reasoning is constrained to a single paragraph, or even part of it. Extended work was done to meet practical demand, such as DrQA~\cite{drqa} answering a SQuAD question based on the whole Wikipedia instead of single paragraph. Besides, latest large-scale datasets, e.g. TriviaQA~\cite{triviaqa} and NarrativeQA~\cite{narrativeqa}, address this limitation by introducing multiple documents, ensuring reasoning cannot be done within local information.
Although those datasets are fairly challenging, reasoning are within one document. 

In many scenarios, we need to comprehend the relationships of entities across documents before answering questions. 
Therefore, reading comprehension tasks with multiple hops were proposed to make it available for machine to tackle such problems, e.g. QAngaroo task~\cite{qangaroo}.
Each sample in QAngaroo contains multiple supporting documents, and the goal is selecting the correct answer from a set of candidates for a query. 
Most queries cannot be answered depending on a single document, and multi-step reasoning chains across documents are needed.
Therefore, it is possible that understanding a part of paragraphs loses effectiveness for multi-hop inference, which posts a huge challenge for previous models.
Some baseline models, e.g. BiDAF~\cite{bidaf} and FastQA~\cite{fastqa}, which are popular for single-document QA, suffer dramatical accuracy decline in this task.

In this paper, 
we propose a new graph-based QA model, named Bi-directional Attention Entity Graph convolutional network (BAG).
Documents are transformed into a graph in which nodes are entities and edges are relationships between them.
The graph is then imported into graph convolutional networks (GCNs) to learn relation-aware representation of nodes. 
Furthermore, we introduce a new bi-directional attention between the graph and a query with multi-level features to derive the mutual information for final prediction.

Experimental results demonstrate that BAG achieves state-of-the-art performance on the WIKIHOP dataset. Ablation test also shows BAG benefits from the bi-directional attention, multi-level features and graph convolutional networks.

Our contributions can be summarized as:
\vspace{-0.2cm}
\begin{itemize}
\setlength{\itemsep}{-0.15cm}
\item Applying a bi-directional attention between graphs and queries to learn query-aware representation for reading comprehension.
\item Multi-level features are involved to gain comprehensive relationship representation for graph nodes during processing of GCNs.
\end{itemize}

\section{Related Work}


Recently coreference and graph-based models are studied for multi-hop QA~\cite{corefgru, santoro2017simple}.
Coref-GRU~\cite{corefgru} uses coreferences among tokens in documents. However, it is still limited by the long-distance relation propagation capability of RNNs. 
Besides, graph is proved to be an efficient way to represent complex relationships among objects and derive relational information~\cite{santoro2017simple}. 
MHQA-GRN~\cite{mhqa} and Entity-GCN~\cite{entitygcn} construct entity graphs based on documents to learn more compact representation for multi-hop reasoning and derive answers from graph networks. However, both of them care less about input features and the attention between queries and graph nodes. 


Attention has been proven to be an essential mechanism to promote the performance of NLP tasks in previous work~\cite{seq2seqattention, memnet}. In addition, bi-directional attention~\cite{bidaf} shows its superiority to vanilla mutual attention because it provides complementary information to each other for both contexts and queries. However, little work exploits the attention between graphs and queries.

\begin{figure*}[ht]
    \setlength{\belowcaptionskip}{-0.4cm}
    \setlength{\abovecaptionskip}{0.2cm}
    \centering
    \includegraphics[width=1\textwidth]{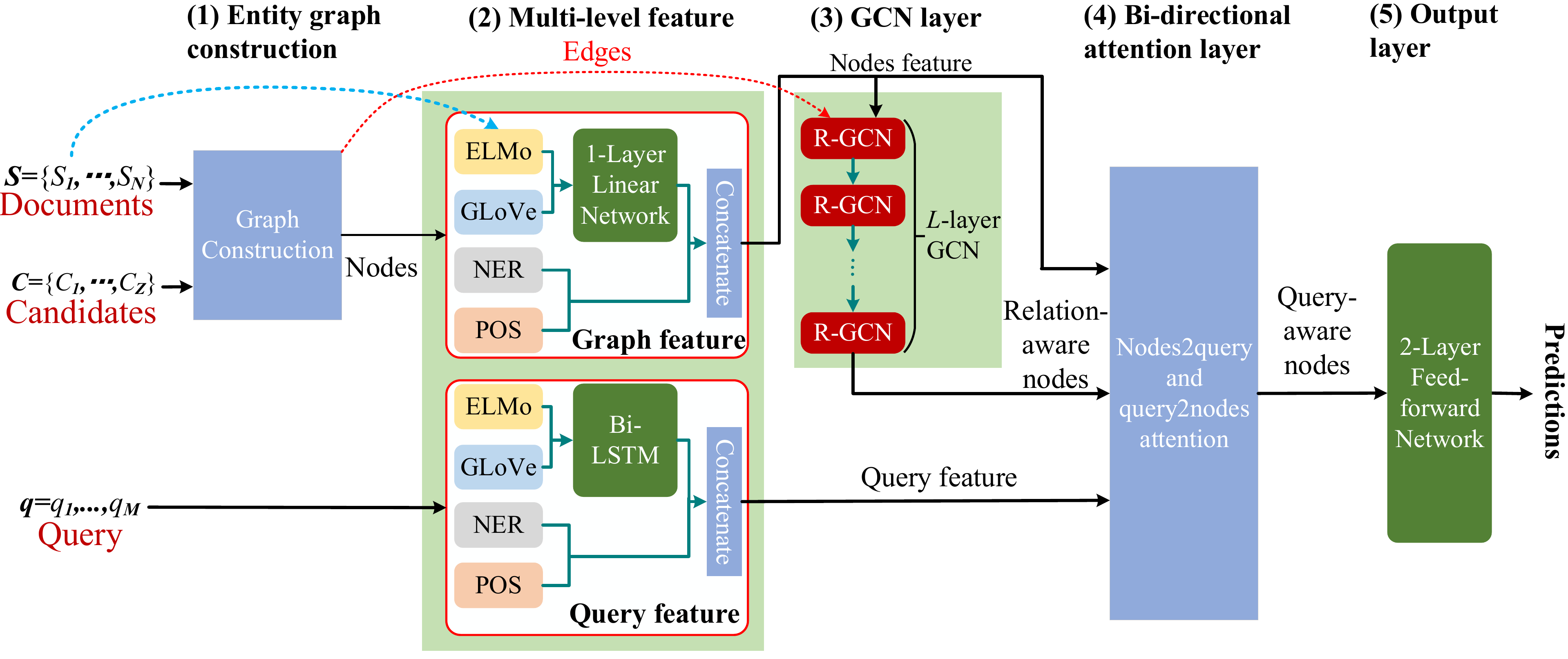}
    \caption{Framework of BAG model.}
    \label{fig:framework}
\end{figure*}

\section{BAG Model}
We first formally define the multiple-hop QA task, taking  QAngaroo~\cite{qangaroo} WIKIHOP data as an example, 
There is a set \bm{$S$} containing $N$ supporting documents, a query \bm{$q$} with $M$ tokens and a set of answer candidates \bm{$C$}. Our goal is to find the correct answer index $a$. 
Giving a triple-style query $\bm{q}=(\textit{country, kepahiang})$, it means \textit{which country does kepahiang belongs to}.
Then answer candidates are provided, e.g. $C=\{\textit{Indonesia}, \textit{Malaysia}\}$. 
There are multiple supporting documents but not all of them are related to reasoning, e.g. \textit{Kephiang is a regency in Bengkulu}, \textit{Bengkulu is one of provinces of Indonesia}, \textit{Jambi is a province of Indonesia}. We can derive the correct candidate is \textit{Indonesia}, i.e. $a=0$, based on reasoning hops in former two documents.



We show the proposed BAG model in Figure~\ref{fig:framework}. It contains five modules: (1) entity graph construction, (2) multi-level feature layer, (3) GCN layer, (4) bi-directional attention and (5) output layer.

\subsection{Entity Graph Construction}
We construct an entity graph based on Entity-GCN~\cite{entitygcn}, which means all mentions of candidates found out in documents are used as nodes in the graph. 
Undirected edges are defined according to positional properties of every node pair. There are two kinds of edges included: 1) cross-document edge, for every node pair with the same entity string located in different documents; 2) within-document edge, for every node pair located in the same document.

Nodes in an entity graph can be found out via simple string matching. This approach can simplify calculation as well as make sure all relevant entities are included in the graph. Picked out along possible reasoning chains during dataset generating~\cite{qangaroo}, answer candidates have contained all related entities for answering. Finally, We can obtain a set of $T$ nodes ${\{n_i\},1 \le i \le T}$ and corresponding edges among these nodes via above procedures.



\subsection{Multi-level Features}

We represent both nodes and queries using multi-level features as shown in Figure~\ref{fig:framework}(2). 
We first use pretrained word embeddings to represent tokens, such as GLoVe~\cite{glove} because nodes and queries are composed of tokens.
Then contextual-level feature is used to offset the deficiency of GLoVe. Note that only part of tokens are remained during graph construction because we only extract entities as nodes. Thus contextual information around these entities in original document becomes essential for indicating relations between tokens and we use higher-level information for nodes except for token-level feature.

We use ELMo~\cite{elmo} as contextualized word representations, modeling both complex word characteristics and contextual linguistic conditions. 
It should be noted that ELMo features for nodes are calculated based on original documents, then truncated according to the position indices of nodes.
Token-level and context-level features will be concatenated and encoded to make a further comprehension. Since a node may contain more than one token, we average features among tokens to generate a feature vector for each node before encoding it. It will be transformed into the encoded node feature via a 1-layer linear network. 

Different from nodes, we represent a query by directly using a bidirectional LSTM (Bi-LSTM) whose output in each step is used as encoded query features. And both linear network and LSTM have the same output dimension ${\hat d}$.

In addition, we add two manual features to reflect the semantic properties of tokens, which are named-entity recognition (NER) and part-of-speech (POS). The complete feature $f_n \in {\mathbb R}^{T \times d}$, $f_q \in {\mathbb R}^{M \times d}$ for both nodes and queries will be the concatenation of corresponding encoded features, NER embedding and POS embedding, where $d= \hat{ d}+{d_{POS}}+{d_{NER}}$.

\subsection{GCN Layer}
In order to realize multi-hop reasoning, we use a Relational Graph Convolutional Network (R-GCN)~\cite{rgcn} that can propagate message across different entity nodes in graphs and generate transformed representation of original ones. The R-GCN is employed to handle high-relational data characteristics and make use of different edge types. At $l$th layer, given the hidden state $h_i^l \in \mathbb{R}^d$ of node $i$, the hidden states $h_j^l \in \mathbb{R}^d,j \in \{N_i\}$ and relations $R_{{N_i}}$ of all its neighbors ($d$ is the hidden state dimension), the hidden state in the next layer can be obtained via
\begin{equation}
    h_i^{l + 1} = \sigma (\sum\limits_{r \in {R_{{N_i}}}} {\sum\limits_{j \in {N_i}} {\frac{1}{{{c_{i,r}}}}W_r^lh_j^l + W_0^lh_i^l)} }, 
\end{equation}
where $c_{i,r}$ is a normalization constant $\left| {N_i^{}} \right|$, $W_r^l \in \mathbb{R}^{d \times d}$ stands a relation-specific weight matrix and $W_0^l \in \mathbb{R}^{d \times d}$ stands a general weight.

Similar to Entity-GCN~\cite{entitygcn}, we apply a gate on update vector $u_i^l$ and hidden state $h_i^l$ of current node by a linear transformation ${f_s}$,
\begin{equation}
    w_i^l = \sigma ({f_s}({\rm{concat}}(u_i^l,h_i^l)), 
\end{equation}
in which $u_i^l$ can be obtained via (1) without sigmoid function. Then it will be used for updating weights for the hidden state $h_i^{l+1}$ of the same node in next layer,
\begin{equation}
    h_i^{l + 1} = w_i^l \odot \tanh (u_i^l) + (1 - w_i^l) \odot h_i^l.
\end{equation}

We stack such networks for $L$ layers in which all parameters are shared.  The information of each node will be propagated up to $L$-node distance away, generating $L$-hop-reasoning relation-aware representation of nodes. The initial input will be mutli-level nodes features ${{\bf{f}}_{\bf{n}}} = \{ {f_{n_i}}\},0 \le i \le T$ and edges ${\bf{e}} = \{ {e_{ij}}\}$ in the graph.

\subsection{Bi-directional Attention Between a Graph and a Query}

Bi-directional attention is responsible for generating the mutual information between a graph and a query.
In BiDAF~\cite{bidaf}, attention is applied to sequence data in QA tasks such as supporting texts. However, we also find it works well between graph nodes and queries. It generates query-aware node representations that can provide more reasoning information for prediction. 

What differs in BAG is that attention is applied 
for graphs as shown in Figure~\ref{fig:framework}(4).
The similarity matrix $\bf{S} \in {\mathbb{R}^{T \times M}}$ is calculated via
\begin{equation}
    {\bf{S}}{\rm{ = av}}{{\rm{g}}_{ - 1}}{f_a}({\rm{concat}}(h_n,f_q,h_n \circ f_q)), \
\end{equation}
in which ${h}_n \in {\mathbb{R}^{T \times d}}$ is all node representations obtained from the last GCN layer, ${f_q} \in {\mathbb{R}^{M \times d}}$ is the query feature matrix after encoding, $d$ is the dimension for both query feature and transformed node representation, $f_a$ is a linear transformation, $\rm{av{g_{ - 1}}}$ stands for the average operation in last dimension, and $\circ$ is element-wise multiplication.

Unlike the context-to-query attention in BiDAF, we introduce a node-to-query attention ${\widetilde a_{n2q}} \in {\mathbb{R}^{T \times d}}$, which signifies the query tokens that have the highest relevancy for each node using 
\begin{equation}
    {\widetilde a_{n2q}} = {\rm{softma}}{{\rm{x}}_{{\rm{col}}}}{\rm{(}}{\bf S}) \cdot f_q,\
\end{equation}
where ${\rm{softma}}{{\rm{x}}_{{\rm{col}}}}$ means performing softmax function across the column, and $\cdot$ stands for matrix multiplication. 

At the same time, we also design query-to-node attention ${\widetilde a_{q2n}} \in {\mathbb R^{M \times d}}$ which signifies the nodes that are most related to each token in the query via
\begin{equation}
    {\widetilde a_{q2n}} = {{\rm dup}({\rm{softmax(ma}}{{\rm{x}}_{{\rm{col}}}}{\rm{(}}{\bf S})))}^\top \cdot  f_n\, ,
\end{equation}
in which ${\rm{ma}}{{\rm{x}}_{{\rm{col}}}}$ is the maximum function applied on across column of a matrix, which will transform $\bf S$ into $\mathbb R^{1 \times M}$. Then function ${\rm dup}$ will duplicate it for $T$ times into shape $\mathbb R^{T \times M}$. $f_n \in {\mathbb R}^{T \times d} $ is the original node feature before GCN layer.

Our bi-directional attention layer is the concatenation of the original nodes feature, nodes-to-query attention, the element-wise multiplication of nodes feature and nodes-to-query attention, and multiplication of nodes feature and query-to-nodes attention.
It should be noted that the relation-aware nodes representation from GCN layer is just used to calculate the similarity matrix, and original node feature is used in rest calculation to obtain more general complementary information between graph and query. Edges are not taken in account because they are discrete and combined with nodes in GCN layer. The output is defined as 
\begin{equation}
    {\widetilde a = {\rm{concat}}(f_n,{\widetilde a_{n2q}},f_n \circ {\widetilde a_{n2q}},f_n \circ {\widetilde a_{q2n}})}.
\end{equation}

\subsection{Output layer}
A 2-layer fully connect feed-forward network is employed to generate the final prediction, with $\tanh$ as the activation function in each layer. Softmax will be applied among the output. It uses query-aware representation of nodes from the attention layer as input, and its output is regarded as the probability of each node becoming answer. Since each candidate may appear several times in the graph, the probability of each candidate is the sum of all corresponding nodes. The loss function is defined as the cross entropy between the gold answer and its predicted probability.

\section{Experiment}

We used both unmasked and masked versions of the QAngaroo WIKIHOP dataset~\cite{qangaroo} and followed its basic setting, in which masked version used specific tokens such as \textit{\_\_MASK1\_\_} to replace original candidates tokens in documents. 
There are 43,738, 5,129 and 2,451 examples in the training set, the development set and the test set respectively, and test set is not public.

In the implementation\footnote{Source code is available on \url{https://github.com/caoyu1991/BAG}.}, 
we used standard ELMo with a 1024 dimension representation.
Besides, 300-dimension GLoVe pre-trained embeddings from 840B Web crawl data were used as token-level features. 
We used spaCy to provide additional 8-dimension NER and POS features.
The dimension of the 1-layer linear network for nodes in multi-level feature module was 512 with $\mathrm{tanh}$ as activation function. 
A 2-layer Bi-LSTM was employed for queries whose hidden state size is 256. 
Then the feature dimension is $d=512+8+8=528$.
The GCN layer number $L$ was set as 5. And the unit number of intermediate layers in output layer was $256$.

In addition, the number of nodes and the query length were truncated as 500 and 25 respectively for normalized computation. Dropout with rate $0.2$ was applied before GCN layer. Adam optimizer is employed with initial learning rate $2\times10^{-4}$, which will be halved for every $5$ epochs, With batch size $32$. It took about $14$ hours for 50-epoch training on two GTX1080Ti GPUs using pre-built and pre-processed graph data generated from original corpus, which can significantly decrease the training time.


We consider the following baseline models: FastQA~\cite{fastqa}, BiDAF~\cite{bidaf}, Coref-GRU~\cite{corefgru}, MHQA-GRN~\cite{mhqa}, Entity-GCN~\cite{entitygcn}. Former three models are RNN-based models, while coreference relationship is involved in Coref-GRU. The last two models are graph-based models specially designed for multi-hop QA tasks.


As shown in Table~\ref{accuracy}, we collected three kinds of results. The dev and test results stand for the original validation and test sets, noting that the test set is not public. Thus, in addition, we divide the original validation set into two parts evenly, one as a split validation set for tuning model and the other one as a split test set. The test$^1$ results are for the split test set. And results of masked dev and test$^1$ are also available in this table.

Our BAG model achieves state-of-the art performance on both unmasked and masked data\footnote{The paper was written on early Dec. 2018, during that time Entity-GCN is the best public model, and only one anonymous model is better than it.}, with accuracy 69.0\% on the test set, which is 1.4\% higher in value than previous best model Entity-GCN. It is significant superior than FastQA and BiDAF due to leveraging of relationship information given by the graph and abandoning some distracting context in multiple documents. Although Coref-GRU extends GRU with coreference relationships, it is still not enough for multi-hop because hop relationships are not limited to coreference, entities with the same strings also existed across documents which can be used for reasoning. Both MHQA-GRN and Entity-GCN utilize graph networks to resolve relations among entities in documents. However, the lack of attention and complementary features limits their performance. Therefore our BAG model achieves the best performance under all data configurations. It is noticed that BAG only gets a small promotion on masked data. We argue that the reason is the attention between masks and queries generating less useful information compared to unmasked ones.

Moreover, ablation experimental results on unmasked version of the WIKIHOP dev set are given in Table~\ref{ablation}. Once we remove the bi-directional attention and put the concatenation of nodes and queries directly into the output layer, it shows significant performance drop with more than $3\%$, proving the necessity of attention for reasoning in multi-hop QA. If we use linear-transformation-based single attention $a=h_n{\bf W_a}f_q$ given in ~\cite{singleattention} instead of our bi-directional attention, the accuracy drops with $2\%$, which means attention bi-directionality also contributes to the performance improvement. The similar condition will appear if we remove GCN, but use raw nodes as input for the attention layer. 

In addition, if edge types are no longer considered, which makes R-GCN degraded to vanilla GCN, noticeable accuracy loss about $2\%$ appears. The absence of multi-level features will also cause degradation. The removal of semantic-level features causes slight decline on the performance, including NER and POS features. Further removal of ELMo feature will causes a dramatical drop, which reflects the insufficiency of only using word embeddings as features for nodes and that contextual information is very important.

\begin{table}[t!]
\setlength{\abovecaptionskip}{-0cm}
\setlength{\belowcaptionskip}{-0cm}
\begin{center}
\begin{threeparttable}
\begin{tabular}{|l|c|c|c|c|c|}
\hline \multirow{2} * {\bf Models} & \multicolumn{3}{c|}{\bf Unmasked} & \multicolumn{2}{c|}{\bf Masked} \\ 
\cline{2-6}
& dev & test & test$^1$ & dev &test$^1$ \\ \hline
FastQA & 27.2$^*$ & - & 38.5 & 38.0$^*$ & 48.3\\
BiDAF & 49.7$^*$ & - & 45.2 & 59.8$^*$ & 57.5 \\
Coref-GRU$^\dagger$ & 56.0$^*$ & 59.3 & 57.2 & - & - \\
MHQA-GRN$^\ddagger$ & 62.8$^*$ & 65.4 & - & - & -\\
Entity-GCN & 64.8$^*$ & 67.6 & 63.1  & 70.5$^*$ & 68.1\\
BAG & \bf 66.5 & \bf 69.0 & \bf 65.7 & \bf 70.9 & \bf 68.9 \\
\hline
\end{tabular}
\begin{tablenotes}
\footnotesize
\item[*] Means results reported in original papers, others are obtained by running official code.  
\item[$\dagger$] Masked data is not suitable for coreference parsing.
\item[$\ddagger$] Some results are missing due to unavailability of source code.
\end{tablenotes}
\end{threeparttable}
\end{center}
\caption{\label{accuracy} The performance of different models on both masked and unmasked version of WIKIHOP dataset. }
\end{table}

\begin{table}[t!]
\setlength{\abovecaptionskip}{-0cm}
\setlength{\belowcaptionskip}{-0.2cm}
\begin{center}
\begin{tabular}{|l|r|r|}
\hline \bf Models & \bf Unmasked \\ \hline
Without Attention & 63.1  \\
Using Single Attention & 64.5 \\
Without GCN & 63.3 \\
Without edge type & 63.9 \\
Without NER, POS & 66.0 \\
\quad +Without ELMo & 60.5 \\
Full Model & \bf 66.5 \\
\hline
\end{tabular}
\end{center}
\caption{\label{ablation} Ablation test results of BAG model on the unmasked validation set of the WIKIHOP dataset. }
\end{table}

\section{Conclusion}

We propose a Bi-directional Attention entity Graph convolutional network (BAG) for multi-hop reasoning QA tasks. Regarding task characteristics, graph convolutional networks (GCNs) are efficient to handle relationships among entities in documents. We demonstrate that both bi-directional attention between nodes and queries and multi-level features are necessary for such tasks. The former one aims to obtain query-aware node representation for answering, while the latter one provides contextual comprehension of isolated nodes in graphs. Our experimental results not only demonstrate the effectiveness of two proposed modules, but also show BAG achieves state-of-the-art performance on the WIKIHOP dataset.

Our future work will be making use of more complex relations between entities and building graphs in more general way without candidates.

\section*{Acknowledgements}
This work was supported by Australian Research Council Projects under grants FL-170100117, DP-280103424. 



\end{document}